\documentclass{article}

\usepackage{microtype}
\usepackage{graphicx}
\usepackage{subcaption}
\usepackage{booktabs} 
\usepackage{bbm}
\usepackage{hyperref}

 \usepackage[preprint]{icml2026}

\usepackage{amsmath}
\usepackage{amssymb}
\usepackage{mathtools}
\usepackage{amsthm}
\usepackage[capitalize,noabbrev]{cleveref}

\theoremstyle{plain}
\newtheorem{theorem}{Theorem}[section]

\theoremstyle{definition}

\newtheorem{assumption}[theorem]{Assumption}
\theoremstyle{remark}

\usepackage[textsize=tiny]{todonotes}

\icmltitlerunning{Nearest Neighbor for Sequantial Counterfactual Inference}

\begin{document}

\twocolumn[
  \icmltitle{C-kNN-LSH: A Nearest-Neighbor Algorithm for Sequential Counterfactual Inference}



  \icmlsetsymbol{equal}{*}

  \begin{icmlauthorlist}
       \icmlauthor{Jing Wang}{yyy}
   \icmlauthor{Jie Shen}{sch}
   \icmlauthor{Qiaomin Xie}{sch2}
   \icmlauthor{Jeremy C Weiss}{yyy}
  \end{icmlauthorlist}
 \icmlaffiliation{yyy}{National Library of Medicine}
\icmlaffiliation{sch2}{University of Wisconsin-Madison}
\icmlaffiliation{sch}{Stevens Institute of Technology}

\icmlkeywords{Machine Learning, ICML}
\icmlcorrespondingauthor{Jing Wang}{jing.wang20@nih.gov}


  \icmlkeywords{Machine Learning, ICML}

  \vskip 0.3in
]



\printAffiliationsAndNotice{}  

\begin{abstract}
 Estimating causal effects from longitudinal trajectories is central to understanding the progression of complex conditions and optimizing clinical decision-making, such as comorbidities and long COVID recovery. We introduce \emph{C-kNN--LSH}, a nearest-neighbor framework for sequential causal inference designed to handle such high-dimensional, confounded situations. By utilizing locality-sensitive hashing, we efficiently identify ``clinical twins'' with similar covariate histories, enabling local estimation of conditional treatment effects across evolving disease states. To mitigate bias from irregular sampling and shifting patient recovery profiles, we integrate neighborhood estimator with a doubly-robust correction. 
 Theoretical analysis guarantees our estimator is consistent and second-order robust to nuisance error. 
 Evaluated on a real-world Long COVID cohort with 13,511 participants, \emph{C-kNN-LSH} demonstrates superior performance in capturing recovery heterogeneity and estimating policy values compared to existing baselines.
\end{abstract}

\section{Introduction}
Long COVID, or Post-Acute Sequelae of SARS-CoV-2 infection (PASC), is a chronic condition characterized by persistent and fluctuating symptoms that can last for months or years following acute infection \cite{wang2026immunological,wang2025active}. Clinical severity is commonly summarized by multiple symptom, such as the PASC score, which evolve irregularly over time and exhibit substantial heterogeneity across patients. A central scientific question is how medical interventions, most notably COVID-19 vaccination, affect the trajectory of Long COVID severity, and how this effect depends on vaccination timing, patient history, and latent physiological state.

Specifically, two patients are with similar early PASC trajectories and comorbidity profiles. One receives a vaccine dose shortly after symptom onset and subsequently experiences a gradual reduction in symptom severity; another receives the same dose much later and shows no improvement or even transient worsening. Understanding such phenomena requires reasoning about counterfactual longitudinal trajectories: how a patient’s future severity would have evolved had vaccination occurred at a different time, or not at all. This problem is inherently sequential, time-dependent, and heterogeneous, and cannot be reduced to a static average treatment effect.

Formally, we observe a cohort of $N$ individuals followed over time. For individual $i$ at time $t$, let $	\mathcal{H}_{i,t} $
denote the observed history, where includes demographics, comorbidities, and high-frequency wearable measurements.   encodes vaccine doses administered at time  . s the observed PASC severity outcome. Let  $Y_{i,t}(a)$ denote the potential outcome under treatment $a$ at time $t$. Our goal is to estimate counterfactual conditional means $\theta_{i,t}(a)$
and, more generally, counterfactual severity trajectories under alternative vaccination schedules.

Classical causal inference approaches often frame such problems in terms of structural causal models (SCMs) and directed acyclic graphs. While powerful, these frameworks require either strong parametric assumptions or explicit recovery of causal structure, tasks that are particularly challenging in biomedical settings with high-dimensional, irregularly sampled longitudinal data. Recent work has argued that, in many applied domains, explicit causal graph recovery is not necessary for answering well-posed causal questions \cite{mlodozeniecposition}. Instead, sufficiently expressive probabilistic models of the longitudinal data-generating process, combined with correct temporal ordering and conditional independence assumptions, are already sufficient to support valid counterfactual and interventional reasoning. This perspective is especially relevant in medicine, where interventions are temporally constrained, covariates are richly observed, and causal questions concern trajectory contrasts rather than abstract graph identification.

In this work, we adopt this viewpoint. Instead of learning a causal graph governing Long covid progression,  we seek to estimate counterfactual outcome trajectories by modeling the conditional distribution $p(Y_{i,t}|H_{i,t},A_{i,t})$ under standard assumptions of sequential consistency, positivity, and ignorability. Our focus is on how probabilistic modeling, when combined with principled nonparametric inference, suffices for causal reasoning in complex longitudinal settings.

\paragraph{Challenges} Despite this conceptual simplification, estimating $\theta_{i,t}(a)$ in practice remains difficult. Three challenges are particularly acute in Long covid studies:
\begin{itemize}
	\item High-dimensional histories. Patient histories include hundreds of clinical variables and dense wearable time series, rendering direct nonparametric matching infeasible. For example, there 500 pages of recover participants.
	\item Time-varying confounding. Vaccination timing depends on evolving health status, which itself predicts future outcomes. such as drug use and evolvement of disease.
	\item Heterogeneity and regime shifts. Empirically, patients exhibit distinct severity regimes, including stable low severity, vaccine-responsive trajectories, and persistent high severity over long follow-up horizons. 
\end{itemize}
Recent theoretical work in statistics has shown that nearest-neighbor matchmaking can yield consistent counterfactual estimators in sequential experiments when histories admit a low-dimensional latent structure. However, these results typically assume access to a sufficient latent state or operate in settings where the state dimension is modest, assumptions violated by modern biomedical datasets.

Hence, we propose a new algorithmic framework that extends nearest-neighbor counterfactual inference to high-dimensional longitudinal data by integrating three components:
\begin{itemize}
	\item Latent history compression. A variational autoencoder (VAE) learns a low-dimensional latent representation of patient history. The representation is trained to preserve outcome-relevant information while acting as an approximate sufficient statistic for treatment assignment.
	\item Local nonparametric inference. Counterfactual outcomes are estimated via k-nearest-neighbor matching in the learned latent space, enabling individualized, trajectory-level inference.
	\item Doubly robust debiasing. We incorporate outcome regression and propensity modeling to obtain estimators that are consistent if either nuisance component is correctly specified.
\end{itemize}
Crucially, the VAE is not used as a black-box predictor. Instead, it serves as a probabilistic state abstraction, enabling classical nonparametric estimators, originally developed under low-dimensional assumptions to operate in complex longitudinal environments. We show that this combination yields consistent counterfactual estimators under approximate latent sufficiency, a setting not covered by existing theoretical work. We apply our method to a large longitudinal Long COVID cohort comprising over 13,511 participants followed for up to 700 days. In addition to clinical covariates and vaccination records, the dataset includes continuous wearable measurements capturing sleep, activity, and physiological signals. These measurements act as high-frequency proxies for latent health state, reducing unobserved confounding and strengthening the credibility of conditional counterfactual comparisons. 

\section{Problem Setup}

We consider a longitudinal observational setting with $N$ units, where each 
individual $i \in [N]$ is tracked over a discrete temporal horizon $t \in [T]$. 
At each time step $t$, we record a feature vector $X_{i,t} \in \mathcal{X} 
\subseteq \mathbb{R}^d$ comprising static comorbidities and high-dimensional 
physiological time-series derived from wearable sensors. Let $A_{i,t} \in 
\mathcal{A}$ denote the intervention (e.g., vaccine dose administration) and 
$Y_{i,t} \in \mathbb{R}$ denote the observed clinical outcome.

We define the observed history up to the moment of treatment assignment at 
time $t$ as the filtration:
\begin{equation}
	\mathcal{H}_{i,t} \doteq \{X_{i,1:t}, A_{i,1:t-1}, Y_{i,1:t-1}\}
\end{equation}

Following the potential outcomes framework, for each action $a \in \mathcal{A}$, 
let $Y_{i,t}(a)$ denote the counterfactual outcome. Our primary estimate is the 
conditional counterfactual mean:
\begin{equation}
	\theta_{i,t}(a) = \mathbb{E}[Y_{i,t}(a) \mid \mathcal{H}_{i,t}]
\end{equation}

We assume the standard longitudinal causal conditions of sequential 
positivity, ignorability, and consistency hold $\forall t, i$.

\begin{assumption}(Consistency). If $A_{i,t}=a$, then $Y_{i,t}=Y_{i,t}(a)$.
\end{assumption}
It asserts a deterministic mapping from the potential outcome space to the observed data manifold.
\begin{assumption}(Sequential ignorability). For all $a\in \mathcal{A}$, $Y_{i,t}(a) \perp\!\!\!\perp A_{i,t} \mid H_{i,t}$.
\end{assumption}
This assumption asserts that the set of observed covariates in $H_{i,t}$ is sufficient to block all ``back-door'' paths between the action $A_{i,t}$ and the future potential outcome. That is, in Directed Acyclic Graphs (DAGs), there exists no unobserved variable $U$ such that $A_{i,t} \leftarrow U \to Y_{i,t}$.
\begin{assumption}(Positivity). There exists $\delta>0$ such that
	\begin{align}
		\delta \leq P(A_{i,t} = a \mid H_{i,t}) \leq 1 - \delta \quad \text{for all } a, i, t.
	\end{align}
\end{assumption}
Given the assumptions above, we let $\Theta$ represents functional mappings that governs the potential outcomes. This work consider the contextual bandit, $\theta_{i,t}(a)$ might represent the potential outcome for unit $i$ at time $t$ under action $a$. The goal is to estimate the full collection of causal parameters such that
\begin{align}
	\Theta = \{ \theta_{i,t}(a) : i \leq N, t \leq T_i, a \in \mathcal{A} \},
\end{align}
which enables individualized counterfactual trajectory inference and downstream interventional contrasts.

\subsection{LLM-Augmented History Compression}
Direct estimation of the causal parameter $\theta_{i,t}(a)$ from raw history $H_{i,t}$ is hindered by the curse of dimensionality. We propose a hybrid architecture where a Large Language Model (LLM), denoted as $\Psi_{\text{LLM}}$, acts as a semantic encoder for raw history, followed by a variational compression layer $\Phi_\phi$.

\begin{assumption}[Latent Sufficiency]
	There exists a mapping $\Phi^\star: \mathcal{H} \to \mathbb{R}^d$ such that:
	\begin{equation}
		Y_{i,t}(a) \perp\!\!\perp A_{i,t} \mid Z_{i,t}, \quad Z_{i,t} = \Phi^\star(H_{i,t}).
	\end{equation}
\end{assumption}

We optimize a revised Evidence Lower Bound (ELBO) that balances generative fidelity, reward relevance, and causal invariance:
\begin{equation}
	\label{eq:loss}
	\begin{split}
			\mathcal{L} =& \mathbb{E}_{q_\phi(Z|H)} \left[ \log p_\psi(H|Z) + \lambda \log p_\omega(Y|Z, A) \right] - \\ &\beta \text{KL}(q_\phi \| p) - \alpha I(Z; A)
	\end{split}
\end{equation}
where $q_\phi(Z_{i,t}|H_{i,t})$ is the encoder that maps the high dimensional LLM-embedded history to the parameters of a latent distribution.  $p_\psi(H \mid Z)$ is the decoder that maps the latent state $Z$ back to history space $H$.  $p_\omega(Y \mid Z, A)$ is causal outcome model that predicts reward (outcome).  The KL-divergence term $\text{KL}(q_\phi \| p)$  measures the distance between the learned encoder $q_\phi(Z \mid H)$ and a prior distribution $p(Z)$. It forces the model to map similar histories to overlapping regions in the latent space, which is exactly what makes  kNN possible. $I(Z; A)$ is the mutual information between the latent state and the action. This penalty discourages \textit{policy leakage}, ensuring the representation is balanced across the action space $\mathcal{A}$.  

\subsection{Local Counterfactual Estimation}

Given the latent embeddings $\{\mathcal{Z}_{i,t}\}$, we estimate $\theta_{i,t}(a)$ using a local comparison set $\mathcal{N}_k(i,t,a)$ retrieved via Approximate Nearest Neighbor (ANN) search \cite{wang2022fast}. To achieve sublinear query time, we employ Locality-Sensitive Hashing. First, let's define a family of hash functions $\mathcal{F}$ such that for any two points $a, b \in \mathcal{Z}$:$$P_{h \in \mathcal{F}}[h(a) = h(b)] = f(d(a,b))$$where $f$ is a monotonically decreasing function,  $d: \mathcal{Z} \times \mathcal{Z} \to \mathbb{R}_{\geq 0}$ is the Euclidean distance $\|\cdot\|_2$. In practice, we use $p$-stable distributions to project latent vectors into buckets:$$h(a,\mathbf{w}, c) = \left\lfloor \frac{\mathbf{w} \cdot a + c}{r} \right\rfloor$$where $\mathbf{w}$ is a random vector with entries drawn from a Gaussian $\mathcal{N}(0,1)$, $r$ is the window size, $c \sim \text{Uniform}(0, r)$.

Hence, given the target latent state $Z_{i,t}$ and action $a$, the local comparison set of $k$ nearest neighbors that received treatment $a$ is computed as:
\begin{align}
	\mathcal{N}_k(i,t,a) \approx \arg\min_{\mathcal{S} \subset \mathcal{D}_a, |\mathcal{S}|=k} \sum_{j \in \mathcal{S}} d(Z_{i,t}, Z_j),
\end{align}
where $\mathcal{D}_a$ is the set of historical latent states associated with action $a$. $Z_{i,t}$ is the query state, the latent embedding of individual $j$ at time $t$. $\mathcal{S}$ is picked from the pool of observations where the treatment equal to $a$. The objective function minimizes the aggregated distance to the query state. The search complexity is $\mathcal{O}((NT)^\rho \log NT)$, where $\rho < 1$ depends on the desired approximation ratio. This ensures that the counterfactual estimation $\widehat{\theta}_{i,t}(a)$ remains tractable even as the history buffer grows.

\subsection{Doubly Robust Correction}

 To mitigate bias from the representation $\Phi_\phi$, we apply a Doubly Robust (DR) correction:
\begin{equation}
	\label{eq:db}
	\widehat{\theta}_{i,t}(a) = \frac{1}{k} \sum_{(j,s) \in \mathcal{N}_k} \left[ \widehat{Q} +  \frac{\mathbb{I}(A_{j,s}=a)}{\hat{e}(a|Z_{j,s})} (R_{j,s} - \widehat{Q}) \right]
\end{equation}
where $\widehat{Q}(Z_{j,s}, a)$ is a local outcome model that predicts the outcome $Y$ based on features $Z_{j,s}$ and action $a$. $\hat{e}$ is the propensity score, a model that predicts the probability of receiving a treatment $a$ based on features $Z_{j,s}$.

\subsection{Neural Network Architecture}

In summary, our architecture, Latent Matchmaking Network (LMN), is composed of three primary modules: a semantic encoder, a variational bottleneck, and a multi-headed prediction layer \cite{shen2019robust}. The structural innovation lies in the integration of a Large Language Model (LLM) with a balanced objective that ensures the learned representation is a sufficient statistic for the potential outcomes. Table \ref{tab:network} outlines the structural configuration of the LMN. 

 Existing work in sequential counterfactual inference often relies on simple RNNs or MLPs to process history, which fails to capture the semantic nuances of text.  We utilize a LoRA-finetuned LLM backbone, $\Psi_{\text{LLM}}$, to project $H_{i,t}$ into a high-capacity semantic space. The encoder network $q_\phi$ then compresses this semantic embedding into a low-dimensional Gaussian distribution by
\begin{align}
	q_\phi(Z \mid H) = \mathcal{N}\left(Z; \mu_\phi(\Psi(H)), \text{diag}(\sigma_\phi^2(\Psi(H)))\right)
\end{align}
We define the latent state as $Z = \mu_\phi + \sigma_\phi \odot \epsilon$ where $\epsilon \sim \mathcal{N}(0, I)$. 
$Z$ is fed into the generative head $p_\psi$ to reconstruct history, causal reward head ($p_\omega$) (concatenated with $A$) to predict the severity $Y$, and the information discriminator that approximates the mutual information $I(Z; A)$. The network is optimized end-to-end. 

\begin{table*}[h]
	\centering
	\caption{Model Architecture Components}
	\begin{tabular}{@{}lll@{}}
		\toprule
		\textbf{Module} & \textbf{Input $\to$ Output} & \textbf{Function} \\ \midrule \hline
		Backbone ($\Psi_{\text{LLM}}$) & Transformer (Frozen): $H_{i,t} \to E_{i,t} $ & Semantic feature extraction from raw history. \\ \addlinespace \hline
		Encoder ($q_\phi$) & $E_{i,t} \to \mu, \sigma \to Z_{i,t}$ & Variational compression and stochastic sampling. \\ \addlinespace \hline
		Decoder ($p_\psi$) & $Z_{i,t} \to \widehat{H}_{i,t}$ & Preservation of historical longitudinal structure. \\ \addlinespace \hline
		Outcome Head ($p_\omega$) & $[Z_{i,t}; A_{i,t}] \to \widehat{Y}_{i,t}$ & Task-specific mapping to potential outcome space. \\ \addlinespace \hline
		Discriminator ($D_\eta$) & $Z_{i,t} \to \text{softmax}(\mathcal{A})$ & Adversarial head for Mutual Information penalty. \\ \bottomrule
	\end{tabular}
	\label{tab:network}
\end{table*}

\begin{algorithm}[H]
	\caption{Sequential Latent Matchmaking (SLM) with DR-Correction}
	\label{alg:slm_main}
	\begin{algorithmic}[1]
		\REQUIRE Longitudinal history $\mathcal{D} = \{ (H_{i,t}, A_{i,t}, Y_{i,t}) \}$, treatment $a \in \mathcal{A}$. 
		\STATE \textbf{Step 1: Representation Learning}
		\STATE Initialize LLM backbone $\Psi_{\text{LLM}}$.
		\WHILE{not converged}
		\STATE Sample mini-batch of histories $\{H_{i,t}, A_{i,t}, Y_{i,t}\}$.
		\STATE $E_{i,t} $ \COMMENT{Project history to semantic space}.
		\STATE $Z_{i,t} $\COMMENT{Embedding of space}.
		\STATE Compute total loss $\mathcal{L}$  as Equation \ref{eq:loss}.
		\ENDWHILE
		\STATE \textbf{Step 2: Local Counterfactual Estimation}
		\FORALL{query $Z_{i,t}$}
		\STATE $\mathcal{N}_k(i,t,a) \gets \text{ANN}_k(Z_{i,t})$\COMMENT{Retrieve $k$-nearest neighbors treated with $a$}.
		\STATE Local outcome model $\widehat{Q}(z, a)$ on $\mathcal{N}_k$.
		\STATE Compute propensity $\hat{e}$ on $\mathcal{N}_k$.
		\STATE Compute $\widehat{\theta}_{i,t}(a)$ as Equation \ref{eq:db}.
		\ENDFOR
		\STATE \textbf{RETURN} Counterfactual estimates $\widehat{\Theta} = \{ \widehat{\theta}_{i,t}(a) \}$.
	\end{algorithmic}
\end{algorithm}

\section{Theoretical Discussion}

We provide a theoretical discussion of the proposed Latent Matchmaking Network (LMN) and its counterfactual estimator. Our goal is not to establish sharp minimax rates, but rather to clarify the conditions under which probabilistic representation learning combined with local nonparametric estimation yields valid counterfactual inference in longitudinal settings.

Recall that for each individual $i$ at time $t$, we observe history $H_{i,t}$, action $A_{i,t}$, and outcome $Y_{i,t}$, with potential outcomes $Y_{i,t}(a)$. The estimand of interest is the conditional counterfactual mean
\[
\theta_{i,t}(a) = \mathbb{E}[Y_{i,t}(a) \mid H_{i,t}].
\]

We assume the standard longitudinal causal conditions of consistency, sequential ignorability, and positivity (Assumptions~2.1--2.3).

\paragraph{Latent sufficiency and approximation error.}
Our approach relies on learning a low-dimensional latent representation $Z_{i,t} = \Phi_\phi(H_{i,t})$ such that $Z_{i,t}$ acts as an approximate sufficient statistic for treatment assignment and potential outcomes.

\begin{assumption}(Approximate Latent Sufficiency) \label{as:latent}
	There exists a measurable mapping $\Phi^\star$ such that
	\[
	Y_{i,t}(a) \perp\!\!\!\perp A_{i,t} \mid Z^\star_{i,t} = \Phi^\star(H_{i,t}),
	\]
	and the learned representation satisfies
	\[
	\mathbb{E}\big[ \| \Phi_\phi(H_{i,t}) - Z^\star_{i,t} \|_2 \big] \le \varepsilon_{\text{rep}}.
	\]
\end{assumption}
The quantity $\varepsilon_{\text{rep}}$ captures representation error induced by finite samples, model misspecification, or imperfect optimization of the variational objective. We assume that the latent counterfactual mean is locally smooth. 
Assumption~3.1 should be understood as a representational strengthening of the standard sequential ignorability assumption (Assumption~2.4). While Assumption~2.4 posits that conditioning on the full observed history $H_{i,t}$ suffices to block confounding, Assumption \ref{as:latent} asserts the existence of a lower-dimensional latent summary $Z^\star_{i,t} = \Phi^\star(H_{i,t})$ that preserves this conditional independence. In general, ignorability given $H_{i,t}$ does not imply the existence of such a sufficient representation. Assumption \ref{as:latent} formalizes the additional requirement that confounding information can be compressed without loss. Our method therefore aims to learn an approximation to $Z^\star_{i,t}$, and the resulting estimator incurs an additional bias term when this approximation is imperfect.


\begin{assumption}
	\label{as:lip}
	(Lipschitz continuity)
	For all $a \in \mathcal{A}$, the function $\theta_{i,t}(a)$ is $L$-Lipschitz in $Z^\star_{i,t}$:
	\[
	|\theta_{i,t}(a) - \theta_{j,s}(a)| \le L \| Z^\star_{i,t} - Z^\star_{j,s} \|_2.
	\]
\end{assumption} 

Given a query state $Z_{i,t}$ and action $a$, LMN constructs a $k$-nearest-neighbor set $\mathcal{N}_k(i,t,a)$ in latent space and computes a doubly robust estimator
\[
\widehat{\theta}_{i,t}(a) = \frac{1}{k} \sum_{(j,s) \in \mathcal{N}_k}
\left[
\widehat{Q}(Z_{j,s}, a) +
\frac{\mathbb{I}(A_{j,s} = a)}{\widehat{e}(a \mid Z_{j,s})}
\big(Y_{j,s} - \widehat{Q}(Z_{j,s}, a)\big)
\right].
\]

\paragraph{Consistency}
Under Assumptions~2.1, 2.3, 3.1, and 3.2, if the neighborhood size $k \to \infty$ and $k / (NT) \to 0$ as $NT \to \infty$, and if either the outcome model $\widehat{Q}$ or the propensity model $\widehat{e}$ is consistent, then
\[
\widehat{\theta}_{i,t}(a) \;\xrightarrow{p}\; \theta_{i,t}(a) + O(\varepsilon_{\text{rep}}).
\]

That is, the estimator is consistent up to an additive bias term induced by representation error. When $\varepsilon_{\text{rep}} \to 0$, LMN recovers the true conditional counterfactual mean.

\paragraph{Discussion.}
This result highlights the role of probabilistic representation learning as a state abstraction mechanism instead of graph. It enables classical nonparametric causal estimators to operate in high-dimensional longitudinal environments. Different from recovering an explicit causal graph, LMN relies on temporal ordering, approximate conditional independence, and local smoothness to support valid counterfactual reasoning. This perspective aligns with recent arguments that sufficiently expressive probabilistic models are adequate for causal inference in richly observed sequential data.

\section{Related Works}
Recent literature has pivoted toward combining foundation models with structural causal models (SCMs) and advanced non-parametric estimation techniques for longitudinal data.

Foundation Models and Probabilistic Causality: Recent works explore using Large Language Models (LLMs) not just for text processing, but as zero-shot causal reasoners. While LLMs struggle with numerical precision \cite{wang2025metric,wang2025large}, they are excellent at identifying potential confounders that structured data might miss \cite{smith2025llm, dwivedi2022counterfactual}. A significant trend in this domain is the shift toward viewing causal inference through the lens of pure probabilistic modeling. Recent work argues that any causal question can be answered within the realm of probabilistic modeling without bespoke causal notation, provided the joint distribution of observed and hypothetical ``intervened'' worlds is explicitly specified \cite{mlodozeniec2025position}.

Latent Representation Learning and VAEs: The use of Variational Autoencoders (VAEs) for causal inference is a well-established generative approach to address confounding \cite{louizos2017causalvae}. Recent advancements include the Disentangled Causal VAE (DCVAE), which integrates causal flows into the representation learning process to learn low-dimensional latent factors that exhibit causal relationships \cite{dcvae2024}. Other models utilize double-stacked VAE architectures to separate the underlying factors responsible for treatment assignment and outcome, thereby improving counterfactual estimation accuracy. Our approach, builds on this by leveraging VAEs for latent history compression in sequential settings, specifically to handle the high-dimensional nature of clinical trajectories like Long COVID \cite{thaweethai2025long}. Beyond seminal encoders, recent work revisits balanced representations via domain adaptation and denoising views to reduce distribution shift and improve robustness under confounding~\citep{hassanpour2019rep,yoon2018ganite,schwab2020cfvae,dbrt2024,shen2021sample,shen2022metric}. Our method inherits these benefits while adding approximate nearest neighbors scalability and sequential doubly robust estimation.

Sequential Causal Inference and Optimal Transport: In longitudinal settings, there is a growing trend of using Optimal Transport (OT) to align treatment and control distributions \cite{doe2024variational}. Methods like Universal Neural Optimal Transport (UNOT) utilize discretization-invariant neural operators to predict transport plans across measures of varying sizes, which is crucial for handling irregularly sampled clinical trajectories \cite{unot2025}. Concurrent research also focuses on ``clinical twins'' or nearest-neighbor matching in high-dimensional spaces. Recent work on Dynamic K-Nearest Neighbor Matching (DK-NNM) introduces data-driven strategies to learn sample-specific neighbors, moving beyond uniform K-values to better handle local confounding \cite{xu2025dynamic}.

Efficient Causal Discovery and Estimation: Learning causal graphs from high-dimensional observational streams remains a central challenge, focusing on the theoretical limits of learning from such streams \cite{chen2025efficient}. New lines of work reformulate causal discovery as a continuous constrained optimization problem \cite{brouillard2020differentiable}. Our method, C-kNN-LSH \cite{anonymous2026cknnlsh}, complements these by focusing on efficient, scalable estimation through Locality-Sensitive Hashing (LSH). Furthermore, we maintain statistical rigor by integrating Doubly Robust (DR) corrections \cite{chernozhukov2018double, robins1994estimation}, ensuring our estimators are consistent and second-order robust to nuisance errors in complex disease environments.

\section{Experiments}

\subsection{Dataset}

We evaluate our method on a large longitudinal Long COVID adult cohort consisting of over 13{,}511 participants followed for 6 months \cite{thaweethai2025long}. It belongs to NIH RECOVER program in the goal to understand, treat, and prevent the postacute sequelaeof SARS-CoV-2 infection. For each participant, Post-Acute Sequelae of SARS-CoV-2 (PASC) severity is measured intermittently, with each individual contributing between 4 and 8 observations over time. The outcome variable is a continuous PASC severity score \cite{thaweethai2023development}, where higher values indicate more severe symptoms.

In addition to outcome measurements, the dataset includes dense wearable-derived physiological summaries, such as activity, heart rate, and respiratory statistics, recorded at weekly resolution. Each wearable record is associated with a timestamp and a concept name (e.g., \emph{Average Heart Rate (Weekly, Mean)}, \emph{Steps (Weekly, NumRecords)}), along with a numeric summary value. Vaccination history is recorded as the cumulative number of vaccine doses received prior to each outcome measurement.

\subsection{History Construction}

For each individual \(i\) and outcome time \(t\), we construct a longitudinal history
\[
H_{i,t} = \{ X_{i,s} : t - L \le s \le t \},
\]
where \(X_{i,s}\) denotes wearable-derived measurements and clinical summaries observed at time \(s\), and \(L\) is a fixed look-back window. In all experiments, we use a look-back window of \(L=180\) days.

To prevent temporal leakage, only history entries with timestamps \emph{at or before} the outcome date \(t\) are included; wearable records occurring after \(t\) are discarded. Within the look-back window, history is summarized at multiple temporal scales (7, 30, 90, and 180 days). For each window and each wearable concept, we compute summary statistics including the mean, standard deviation, minimum, maximum, and number of recorded values. These summaries are serialized into a structured textual representation, which serves as input to language-model-based or text-based baselines.

The treatment variable
\[
A_{i,t} \in \{0,1,2,3,4,5,6\}
\]
denotes the cumulative number of vaccine doses received by individual \(i\) prior to time \(t\). This quantity is treated as a multi-valued action and is included both as an explicit treatment variable and, where appropriate, as part of the observed history.

\subsection{Train/Test Splitting Protocol}

To ensure a valid evaluation of counterfactual generalization, we split the data at the \emph{participant level}. Specifically, participants are randomly divided into disjoint training (70\%), validation (10\%), and test (20\%) sets. All outcome measurements for a given participant are assigned to the same split, preventing information leakage across time or individuals.

All models are trained exclusively on the training set. Counterfactual estimates and evaluation metrics are reported on the held-out test set.

\subsection{Baselines}

We compare our proposed \emph{Latent Matchmaking Network (LMN)} against several standard counterfactual estimation baselines:

\begin{itemize}
	\item \textbf{Outcome Regression (OR)}~\citep{robins1986new,hernan2020causal}.  
	A plug-in estimator that fits a regression model for \(\mathbb{E}[Y \mid H, A]\) using TF-IDF features of the constructed history and evaluates this model at counterfactual treatment levels.
	
	\item \textbf{Inverse Propensity Weighting (IPW)}~\citep{rosenbaum1983central,robins2000marginal}.  
	A global estimator that reweights observed outcomes using estimated treatment propensities \(\mathbb{P}(A \mid H)\), without conditioning on local neighborhoods.
	
	\item \textbf{Local AIPW}~\citep{robins1994estimation,chernozhukov2018double}.  
	A doubly robust estimator that combines outcome regression and propensity weighting within local neighborhoods defined by similarity in the history space.
\end{itemize}

All baselines operate on the same leakage-safe history representation and use the same train/test splits.

\subsection{Settings}

Our method, LMN, uses a frozen Qwen3-4B-base language model to encode the textual history representation into a high-dimensional semantic embedding. This embedding is compressed via a variational bottleneck trained with a balanced objective enforcing approximate latent sufficiency and treatment invariance. Counterfactual outcomes are estimated using k-nearest-neighbor matching in the learned latent space, combined with a doubly robust correction. The experiments are conducted on Nvidia PRO 6000.

\subsection{Results}

Table~\ref{tab:counterfactual_comparison} reports the mean estimated counterfactual PASC severity
\[
\widehat{\theta}(a) = \mathbb{E}[\widehat{Y}(a)]
\]
under different numbers of prior vaccine doses \(a\), evaluated on the test set.

\begin{table}[t]
	\centering
	\caption{Mean counterfactual PASC severity $\widehat{\theta}(a)$ under different numbers of prior vaccine doses. Lower values indicate lower symptom severity.}
	\label{tab:counterfactual_comparison}
	\begin{tabular}{c|ccc|c}
		\toprule
		\textbf{$a$} 
		& \textbf{OR} 
		& \textbf{IPW} 
		& \textbf{Local AIPW} 
		& \textbf{LMN (Ours)} \\
		\midrule
		0 & 2.59 & 2.73 & 2.05 & 3.76 \\
		1 & 10.65 & 11.31 & 15.50 & 10.42 \\
		2 & 7.35 & 7.88 & 7.95 & 9.68 \\
		3 & 6.49 & 6.64 & 7.36 & 4.54 \\
		4 & 4.36 & 4.46 & 4.63 & 3.92 \\
		5 & 4.50 & 4.57 & 3.90 & 4.46 \\
		6 & 4.61 & 4.83 & 4.18 & 3.68 \\
		\bottomrule
	\end{tabular}
\end{table}


From Table~\ref{tab:counterfactual_comparison}, all methods reveal pronounced heterogeneity in estimated PASC severity as a function of vaccination exposure. In particular, severity is highest at intermediate vaccine counts ($a=1$ and $a=2$), while lower severity is observed both in the absence of vaccination ($a=0$) and at higher vaccine counts. This non-monotonic pattern highlights the inadequacy of static or monotonic treatment effect assumptions in this longitudinal setting.

Second, outcome regression and inverse propensity weighting produce relatively smooth dose-response curves, reflecting their reliance on global averaging and parametric modeling assumptions. Local AIPW exhibits substantially higher variability, especially at $a=1$, where it predicts markedly higher severity than other methods. This suggests instability in neighborhood construction when matching is performed using surface-level textual similarity in a high-dimensional history space.

In contrast, the proposed Latent Matchmaking Network (LMN) yields a distinct counterfactual trajectory at moderate-to-high vaccine counts. Notably, for $a=3,4,$ and $6$, LMN predicts consistently lower severity than Local AIPW, while remaining comparable to or slightly higher than global estimators at $a=2$ and $a=5$. These differences indicate that the learned latent representation enables more meaningful local comparisons by filtering irrelevant dimensions of the observed history, thereby stabilizing counterfactual estimation in regions of the action space where overlap is limited.

Importantly, LMN achieves these results without explicit causal graph specification. Instead, it relies on probabilistic modeling of longitudinal histories, temporal ordering of observations, and local counterfactual comparisons in a learned latent space. The empirical results therefore support the view that such probabilistic representation-based approaches can be sufficient for causal reasoning in richly observed longitudinal datasets.

\subsection{Causal Effect Estimates Across Concept Sets}
\begin{figure}[t]
	\centering
	\includegraphics[width=\linewidth]{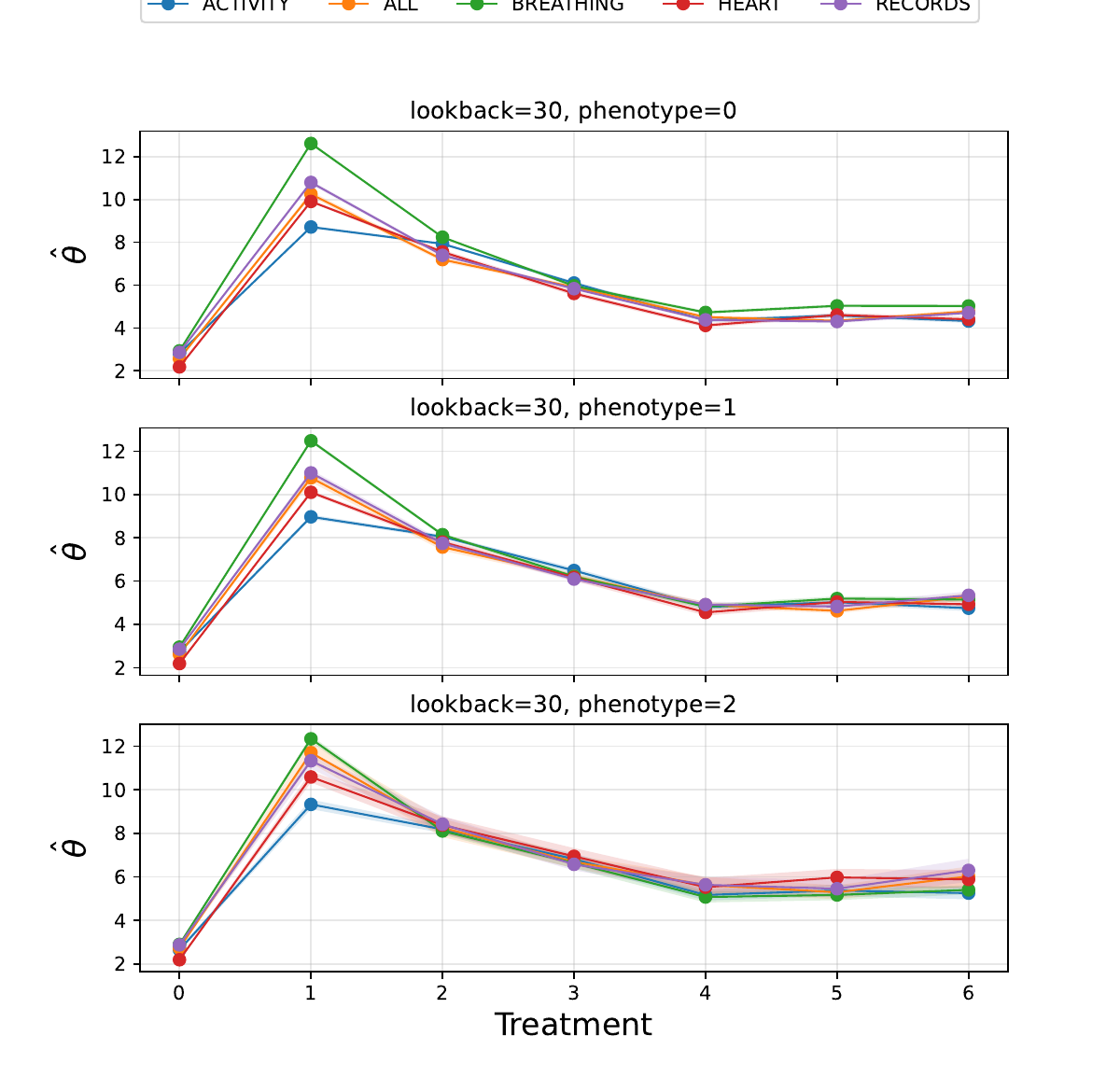}
	\caption{Estimated causal effects $\hat{\theta}(a)$ across treatment actions for three phenotypes under a 30-day history lookback. Each curve corresponds to a different concept set, including the baseline ALL representation. Shaded regions indicate variability across individuals within each phenotype.}
	\label{fig:fig1}
\end{figure}

Figure \ref{fig:fig1} presents the estimated causal effects across treatment actions for three phenotypes under a 30-day history lookback. Each curve corresponds to a different concept set, with ALL denoting the baseline representation using all available concepts. Across all phenotypes, the estimated effect curves exhibit a consistent, non-monotonic structure with a pronounced peak at treatment $a=1$ followed by a gradual decline as the treatment index increases

While the overall shape of the effect curves is similar across concept sets, their magnitudes differ systematically. In particular, concept sets based on ``BREATHING'' and ``ACTIVITY ''features tend to yield larger estimated effects at lower treatment levels, whereas ``HEART'' and ``RECORDS'' concepts produce more conservative estimates. This pattern indicates that representation-level concept selection materially influences causal effect estimation, even when the downstream estimator and population are held fixed. Importantly, these differences are structured rather than noisy, concept-specific curves remain smooth across treatment levels and preserve the same qualitative ordering of actions. This suggests that the learned causal effects are driven by stable signal in the representations rather than estimation artifacts.
\subsection{Heterogeneity by Phenotype}
Figure \ref{fig:fig1} further reveals substantial heterogeneity in treatment effects across phenotypes. Although all phenotypes display a peak effect at treatment $a=1$, the magnitude and decay profile differ meaningfully. For example, phenotype 2 exhibits consistently higher effect estimates across most treatments compared to phenotypes 0 and 1, while phenotype 1 shows a faster attenuation of effects at higher treatment levels

These differences reflect effect modification by phenotype, rather than simple shifts in baseline outcomes. Because individual-level causal effects  $\hat{\theta}(a)$ are estimated prior to aggregation, the observed heterogeneity captures variation in how different subpopulations respond to treatment. This demonstrates the ability of our method to recover heterogeneous treatment effect profiles that would be obscured by population-level averaging.

\subsection{Sensitivity to History Lookback}
\begin{figure}[t]
	\centering
	\includegraphics[width=\linewidth]{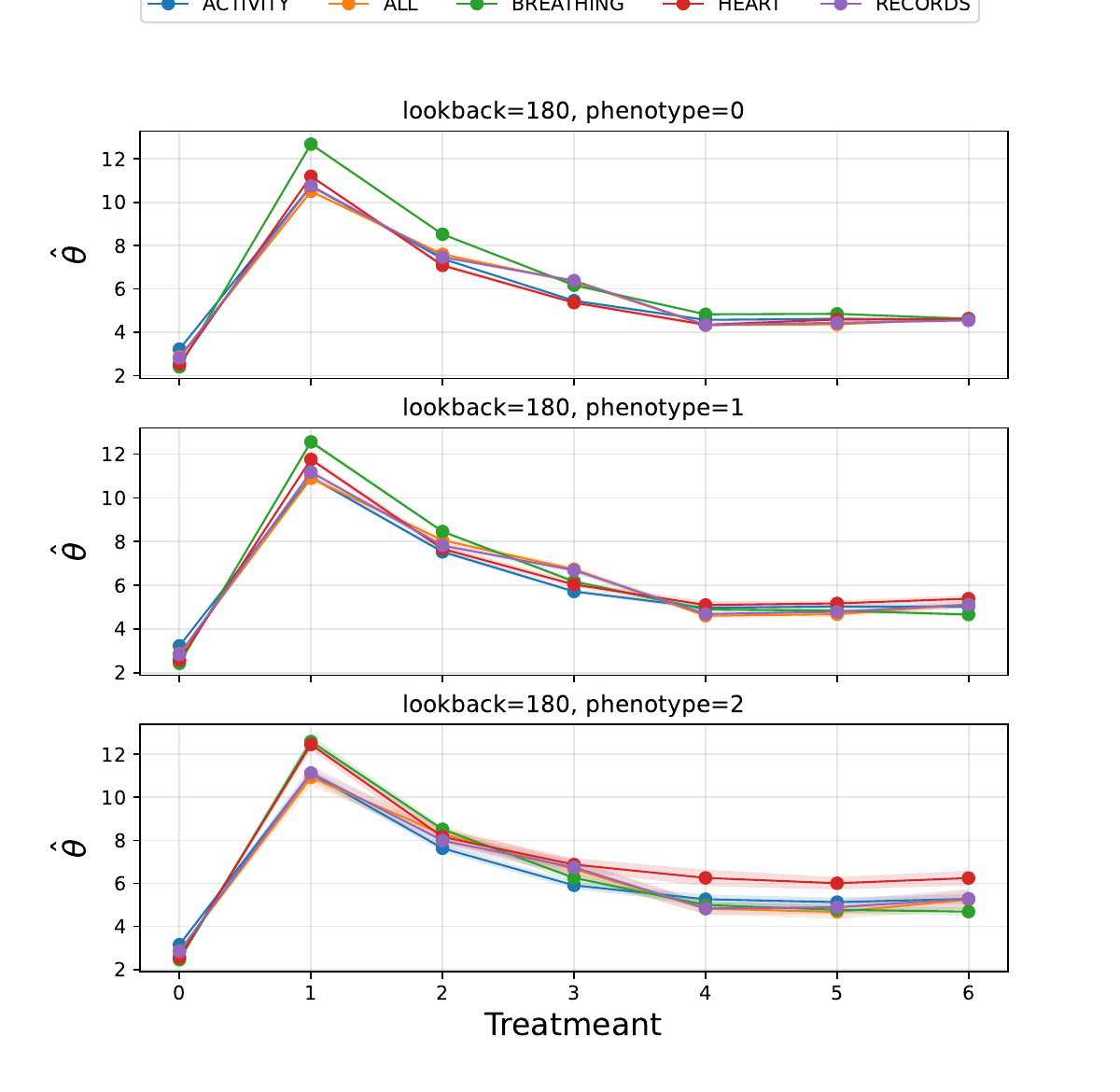}
	\caption{Estimated causal effects $\hat{\theta}(a)$ across treatment actions for three phenotypes under a 180-day history lookback. Each curve corresponds to a different concept set, including the baseline ALL representation. Shaded regions indicate variability across individuals within each phenotype.}
	\label{fig:fig2}
\end{figure}

To assess robustness to temporal context, Figure \ref{fig:fig2} reports analogous results using a longer 180-day lookback window. The estimated causal effect curves remain highly consistent with those obtained under a 30-day lookback, both in shape and relative ordering of treatments and concept sets.

Across all phenotypes, the location of the peak effect and the overall decay pattern are preserved, indicating that the estimated causal relationships are not overly sensitive to the specific choice of history length. Minor changes in magnitude are observed, particularly for higher treatment indices, but these variations are smooth and do not alter the qualitative conclusions.

Together, these results suggest that the proposed estimator produces stable causal effect estimates once sufficient historical information is available.

\subsection{Relative effects compared to the baseline concept set}
\begin{figure}[t]
	\centering
	\includegraphics[width=\linewidth]{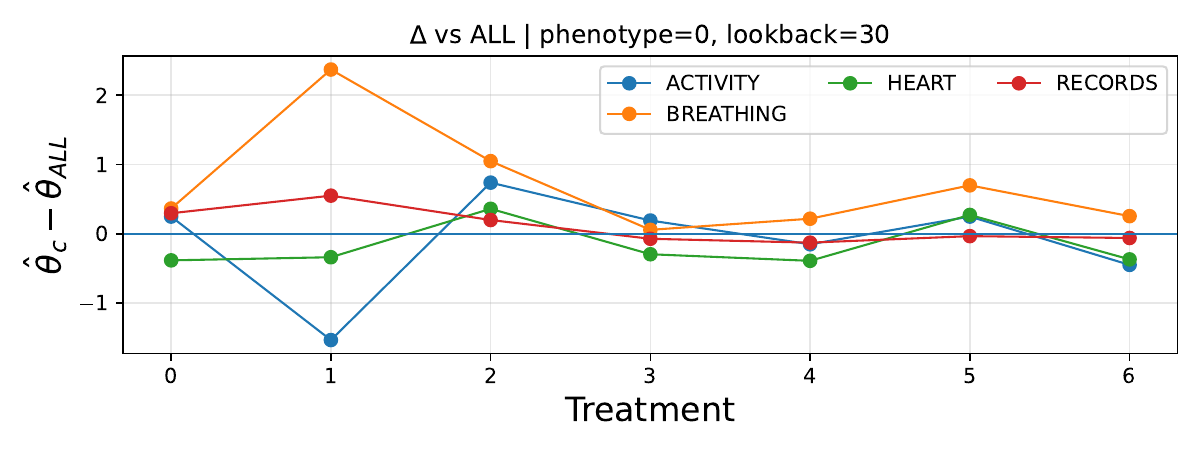}
	\caption{Differences in estimated causal effects relative to the baseline ALL representation ($\hat{\theta}_c-\hat{\theta}_{\text{ALL}}$) for phenotype 0 under a 30-day lookback. Positive values indicate higher estimated effects compared to the baseline.}
	\label{fig:fig3}
\end{figure}

\begin{figure}[t]
	\centering
	\includegraphics[width=\linewidth]{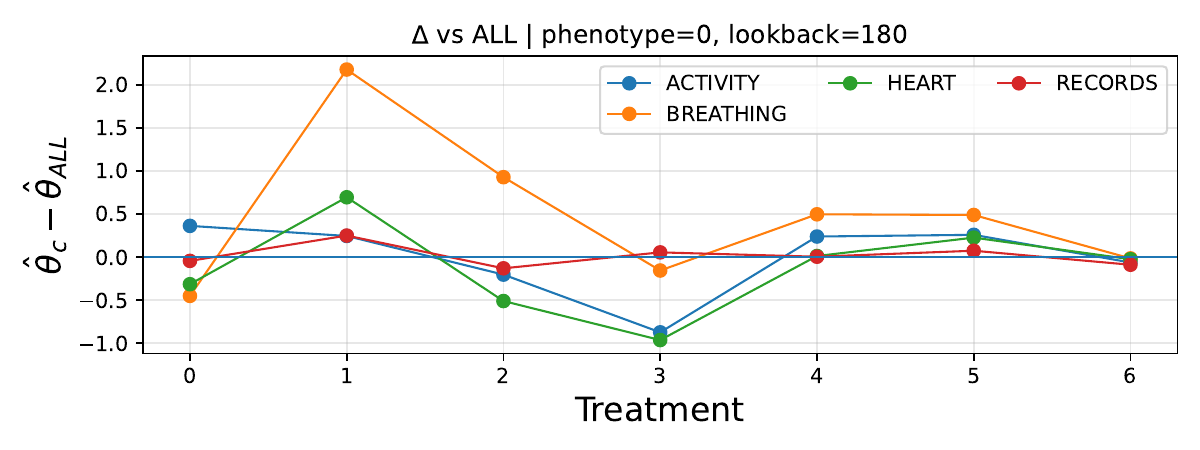}
	\caption{Differences in estimated causal effects relative to the baseline ALL representation ($\hat{\theta}_c-\hat{\theta}_{\text{ALL}}$) for phenotype 0 under a 180-day lookback. Positive values indicate higher estimated effects compared to the baseline.}
	\label{fig:fig4}
\end{figure}

To isolate the contribution of individual concept sets, Figure \ref{fig:fig3} shows the difference in estimated effects relative to the baseline ALL representation for a 30-day lookback, and Figure \ref{fig:fig4} reports the corresponding results for a 180-day lookback . 

Across phenotypes, BREATHING concepts consistently yield positive deviations from the baseline at lower treatment levels, particularly around $a=1$, whereas ACTIVITY and HEART concepts exhibit both positive and negative deviations depending on treatment intensity. RECORDS concepts remain close to the baseline across most settings, indicating limited incremental information beyond the full representation.

Notably, these relative differences are stable across lookback windows: the sign and approximate magnitude of  $\hat{\theta}_c-\hat{\theta}_{\text{ALL}}$ remain similar when increasing the lookback from 30 to 180 days. This stability provides further evidence that the observed concept-level effects reflect meaningful representation differences rather than temporal noise.

Overall, the results demonstrate that our method recovers smooth, interpretable causal effect curves across multiple treatment levels. Treatment effects exhibit clear heterogeneity across phenotypes. The representation choice via concept sets systematically influences causal estimates. The estimated effects are robust to substantial changes in history lookback. These findings support the effectiveness and stability of the proposed representation-aware local causal estimation framework.
\section{Conclusion}
This work introduced C-kNN-LSH, a novel nearest-neighbor framework for sequential counterfactual inference designed to address the challenges of high-dimensional, confounded longitudinal data. By integrating latent history compression via a Variational Autoencoder (VAE) with Locality-Sensitive Hashing (LSH), we enabled efficient identification of ``clinical twins'' for local estimation of conditional treatment effects in complex disease trajectories. Our approach mitigates the curse of dimensionality inherent in raw clinical histories while maintaining statistical rigor through a doubly robust correction that ensures consistency even under imperfectly specified nuisance models.

Empirical evaluation on a real-world Long COVID cohort of 13,511 participants demonstrated that our model, the Latent Matchmaking Network (LMN), effectively captures recovery heterogeneity and provides more stable counterfactual estimates than existing global and local baselines. These results validate the perspective that sufficiently expressive probabilistic representation-based models can support valid causal reasoning in richly observed sequential data without the need for explicit causal graph recovery. Ultimately, C-kNN-LSH provides a scalable and principled tool for understanding the longitudinal impacts of interventions in high-dimensional biomedical settings, offering a path toward more personalized and data-driven clinical decision-making.

\section*{Impact Statement}

%
This paper presents work whose goal is to advance the field of Machine
Learning. There are many potential societal consequences of our work, none
which we feel must be specifically highlighted here.
%

\bibliography{causal}
\bibliographystyle{icml2026}

\end{document}